\documentclass[conference]{IEEEtran}
\IEEEoverridecommandlockouts
\usepackage{cite}
\usepackage{amsmath,amssymb,amsfonts}
\usepackage{algorithmic}
\usepackage[caption=false]{subfig}
\usepackage{placeins}
\usepackage{graphicx}
\graphicspath{ {figures/} }
\usepackage{textcomp}
\usepackage{xcolor}
\usepackage[bookmarks=false]{hyperref}
\usepackage{orcidlink}
\def\BibTeX{{\rm B\kern-.05em{\sc i\kern-.025em b}\kern-.08em
    T\kern-.1667em\lower.7ex\hbox{E}\kern-.125emX}}

\begin{document}
\bstctlcite{bstctl:nodash}
\title{Multivariate Time Series Early Classification Across Channel and Time Dimensions

\thanks{This work has been conducted as part of the Just in
Time Maintenance project funded by the European Fund for
Regional Development.}
}

\author{\IEEEauthorblockN{Leonardos Pantiskas}
\IEEEauthorblockA{\textit{Vrije Universiteit Amsterdam}\\\orcidlink{0000-0002-4898-5334} 0000-0002-4898-5334}
\and
\IEEEauthorblockN{Kees Verstoep}
\IEEEauthorblockA{\textit{Vrije Universiteit Amsterdam}\\\orcidlink{0000-0001-6402-2928} 0000-0001-6402-2928}
\and
\IEEEauthorblockN{Mark Hoogendoorn}
\IEEEauthorblockA{\textit{Vrije Universiteit Amsterdam}\\
\orcidlink{0000-0003-3356-3574} 0000-0003-3356-3574}
\and
\IEEEauthorblockN{Henri Bal}
\IEEEauthorblockA{\textit{Vrije Universiteit Amsterdam}\\
\orcidlink{0000-0001-9827-4461} 0000-0001-9827-4461}
}

\maketitle

\begin{abstract}

Nowadays, the deployment of deep learning models on edge devices for addressing real-world classification problems is becoming more prevalent. Moreover, there is a growing popularity in the approach of early classification, a technique that involves classifying the input data after observing only an early portion of it, aiming to achieve reduced communication and computation requirements, which are crucial parameters in edge intelligence environments. While early classification in the field of time series analysis has been broadly researched, existing solutions for multivariate time series problems primarily focus on early classification along the temporal dimension, treating the multiple input channels in a collective manner. In this study, we propose a more flexible early classification pipeline that offers a more granular consideration of input channels and extends the early classification paradigm to the channel dimension. To implement this method, we utilize reinforcement learning techniques and introduce constraints to ensure the feasibility and practicality of our objective. To validate its effectiveness, we conduct experiments using synthetic data and we also evaluate its performance on real datasets. The comprehensive results from our experiments demonstrate that, for multiple datasets, our method can enhance the early classification paradigm by achieving improved accuracy for equal input utilization.
\end{abstract}

\section{Introduction}

The success of machine and deep learning methods in a multitude of tasks, such as classification of video and time series data, combined with the spread of Internet of Things, with data-rich, sensor-driven applications, has led to a rise of deployment of those solutions on portable computing platforms (e.g. smartphones, drones) close to the data sources, constituting the so-called \textit{edge intelligence} \cite{chenDeepLearningEdge2019,zhouEdgeIntelligencePaving2019}. Although these edge devices are becoming increasingly capable, they are typically more resource constrained than the traditional computational platforms. As a result, in the context of edge intelligence, some metrics that were previously insignificant compared to classification accuracy acquire much higher importance, such as the number of operations or energy required for a given task\cite{chenDeepLearningEdge2019}. 

A popular paradigm to address the above is the cloud-edge collaboration\cite{wangConvergenceEdgeComputing2020}, in which an edge device can offload tasks to a remote, cloud model that is potentially too large or computationally expensive to run on the edge. This approach, however, introduces new significant constraints to the task of classification, namely the bandwidth necessary for data transfer from edge to cloud and the latency required for the classification of a sample. As a result, researchers have developed various approaches that try to tackle these, such as \cite{canel2019fforward,wang2018bandwidth,nigade2020clownfish} in the field of video analytics, which utilize methods to selectively filter the input and transfer to the cloud only the parts most useful for the relevant task. 

Another paradigm to achieve computational efficiency on edge devices is that of early-exit neural networks\cite{han2021dynamic}. This subcategory of dynamic neural networks includes models that can adapt their computation based on each sample and terminate the inference before having observed the full input across the temporal dimension, with examples in domains such as text\cite{zhengjie2017textee} and video classification\cite{fan2018videoee}. When applied to the time series domain, this technique is commonly referred to as \textit{early classification}.

Early classification for time series has been extensively studied and has many benefits and multiple applications\cite{gupta2020ects}. In the healthcare domain, an early classification based on partial time series data can lead to an earlier diagnosis of a disease and a higher chance of successful treatment. In the industrial domain, early classifications based on machinery sensor data can lead to the timely detection of faults and minimize the interruption of the manufacturing processes. Although early classification commonly has an association with savings in time, recent interesting debate\cite{wu2023whenmeaningful,achenchabe2021meaningful} on the matter has given rise to valuable insights and has demonstrated that this is not always the case, as it heavily depends on the use case. For instance, in an industrial environment with high-frequency sensors, where classification of motor status happens within time windows of a few seconds, early classification is unlikely to have significant time savings, and it may even be unnecessary if the decision pipeline (e.g. shutdown in case of a fault) is not automated and there is a human in the loop. However, early classification can still benefit other metrics, which align with the edge intelligence context, such as savings in computation, communication, and energy.

Time series classification has often been tackled with recurrent neural networks (RNNs), so it follows that there are already methods that achieve early classification with RNNs, such as \cite{dennis2018emirnn,rubwurm2019endtoendlearning,kannan2021budgetrnn}. Moreover, there is a much more prolific family of RNN-based solutions\cite{han2021dynamic}, that, although not explicitly targeting early classification in time, use various approaches to try to limit the computation performed per sample, aiming for more efficient RNNs (e.g. \cite{koutnik2014clockwork,campos2018skiprnn}). Although these solutions mainly alter their computation based on the steps across the temporal dimension, they target \textit{univariate time series} problems, i.e. time series with only 1 channel of information. When it comes to \textit{multivariate time series}, with multiple channels as input, these solutions are either not applicable or treat the channel dimension as a singular entity, exhibiting the same behavior per timestep for all channels.

Taking into account all the concepts above, we propose our framework, CHARLEE, for \textbf{Ch}annel-\textbf{A}daptive, \textbf{R}einforcement \textbf{L}earning-based \textbf{E}arly \textbf{E}xiting, which targets multivariate time series specifically and performs early exiting on a more fine-grained manner than the existing early classification paradigm, both across the temporal and channel dimensions. Our framework aims to provide a flexible manner to reduce the data necessary for an inference on a per-sample basis, potentially filtering out different channels across different points in time of the time series input, before propagating it to a final classifier.

A characteristic example of a use case where the usefulness of our framework is demonstrated is the following: In a heavy industry environment, a motor is potentially monitored by current, voltage, and vibration sensors for faults. During a classification of sensor readings, it could be obvious within a few timesteps that a fault is present, but the information about the specific type of fault could lie in the future of only the vibration sensor readings, while the voltage and current sensors would not provide additional useful information. A typical early classification solution could not terminate the inference early without losing accuracy and it would be necessary to continue collecting and processing sensor readings from all sensors until the point where the vibration channel would contain the information necessary to perform the classification correctly. On the other hand, our flexible framework could exit earlier across the uninformative channels, only keeping the vibration sensor readings channel for the classification.

Our framework is a generic solution, but we envision it in the context of edge intelligence, being deployed on an edge device as part of a cloud-edge collaboration scenario. As we mentioned, energy-efficient inference is important in edge intelligence, both because of sustainability purposes, as well as extreme scenarios that require this, such as energy-harvesting systems\cite{gobieski2019intermittent}. Moreover, in a sensor-based edge intelligence environment, a significant factor is the energy consumed by the sensors themselves to collect the necessary data\cite{casare2009energymgmt,kannan2021budgetrnn}. Thus, our fine-grained input-filtering solution, placed on an edge device between the sensors and the cloud, can translate to energy savings across the whole path of sensor data to cloud. For instance, as we discussed in the example above, if our framework decides at a checkpoint to filter out some of the input channels, the sensors generating those channels no longer need to collect or transmit future data, thus conserving data-collecting and radio energy.

Our contributions with CHARLEE are:
\begin{itemize}
    \item We introduce the concept of explicit channel-specific early exiting for multivariate time series classification.
    \item We formulate this concept as a Partially Observable Markov Decision Process and design a reinforcement learning approach to address it.
    \item We make the problem tractable by introducing a set of reasonable constraints while maintaining its usefulness for real-world use cases.
    \item We verify the behavior of our framework employing a synthetic dataset constructed with real dataset signals and we test its performance on 26 real datasets, showing its potential.
\end{itemize}

\section{Related Work}

\subsection{Channel filtering}

An empirical conclusion that has emerged in the field of multivariate time series classification is that not all channels are equally important for a given task\cite{kathirgamanathan2020feature,dhariyal2021fast}. This observation, paired with the fact that the execution time of many classification methods increases significantly for datasets with more channels\cite{ruizGreatMultivariateTime2021}, has led to an effort to develop methods that can select the most useful channels for a given task, such as \cite{yoon2005featuresubset}, \cite{kathirgamanathan2020feature} and \cite{dhariyal2021fast}. Generally, those methods rank the channels depending on their utility to the classification task and filter out the non-helpful ones. However, this is a static process that alters the dataset and is thus orthogonal to our approach, where we assume that the channels may have useful information for the classification at different time points per class, and we do not consider the question of whether a channel can be a priori completely excluded.

\subsection{Adaptive sensor sampling}

A different view of the channel filtering approach, taking into consideration that likely sources of the channels are separate sensors, is adaptive sensor sampling. This sampling can refer to the time dimension, as some works dynamically change the frequency of the incoming sensor readings to conserve energy, such as \cite{alippi2010samplingenergyhungry}. However, there are works where this sampling refers to the sensors themselves, such as \cite{willett2004backcasting} and \cite{gedik2007asap}, which utilize, under different approaches, subsets of all available sensors to sense the environment and conserve energy and communication without sacrificing accuracy. Our work bridges this concept with recent, deep learning-based techniques for time series, and although we do not focus on the hardware and wireless network aspect of sensor subset selection, this sensor selection is a natural consequence of the application of our framework in real-world use cases.

\subsection{Early classification of time series}
The task of early classification in time series is well-studied, with methods used ranging from simple decision trees to complex deep learning models\cite{gupta2020ects}. Despite this multitude of early classification solutions, in this section we focus on methods that are relevant to our approach, either because they address multivariate time series specifically, by adapting their computation depending on the individual channel properties, or because they are based on reinforcement learning.

Firstly, approaches based on shapelets, such as \cite{ghalwash2012early} and \cite{he2015early}, work by finding characteristic signatures for each class in the channels of the multivariate dataset. Although the shapelets can potentially be more interpretable than other solutions, it can be computationally expensive to scan the time series and extract them, and this approach does not scale well as the number of channels and the length of the time series grow.

An interesting approach that represents the multivariate time series as Multivariate Marked Point-Process is presented in \cite{li2014early}. The authors first consider each channel as an event with temporal dynamics that can be extracted, and then the order of patterns across the channels (events) along the time dimension is used as a discriminative cue for classification.

Two model-based approaches that differentiate computation for different input channels are \cite{gupta2020fault} and \cite{gupta2020divide}. The factor that sets the channels apart are their fault status and sampling rate, respectively. For instance, in \cite{gupta2020fault}, the faulty sensor channels are detected at the beginning of each inference and excluded for the rest of the process. Both methods require the construction of multiple classifiers, which can be computationally expensive both during training (for large datasets) and during inference on an edge device, which will have to maintain the classifiers in memory, thus clashing with our stated objective.

There have also been works utilizing deep learning structures, such as \cite{huang2018mtsmtd} and \cite{hsu2019multivariate}, where the authors utilize modules that include convolutional neural networks and long short-term memory networks to extract information from the various channels and subsequently combine them. However, they also do not process the various channels differently.

In the field of reinforcement learning, the authors of \cite{martinez2018deep} and \cite{martinez2020adaptive} approach the problem of early exiting utilizing a deep Q-network\cite{mnih2015human} agent to decide the point of exit and introducing the earliness-accuracy trade-off in its reward function. In a related approach, the authors of \cite{hartvigsen2020recurrent} and \cite{hartvigsen2022stop} use a halting policy network to sample the action of stopping in time (or potentially jumping ahead in the latter work), trained using the REINFORCE algorithm\cite{williams1992simple}. These works do not tackle early exiting in the channel dimension but present useful ideas about the formulation of the task as a partially-observable Markov Decision Process (POMDP).

To the best of our knowledge, no work exists that targets multivariate time series and explicitly applies early exit at different time points for different channels (or groups of channels), allowing complete halting of the computation and even halting of "reading" the input in a fine-grained manner across channels and timesteps.
\begin{figure*}[htbp]
\centering
\includegraphics[width=\linewidth]{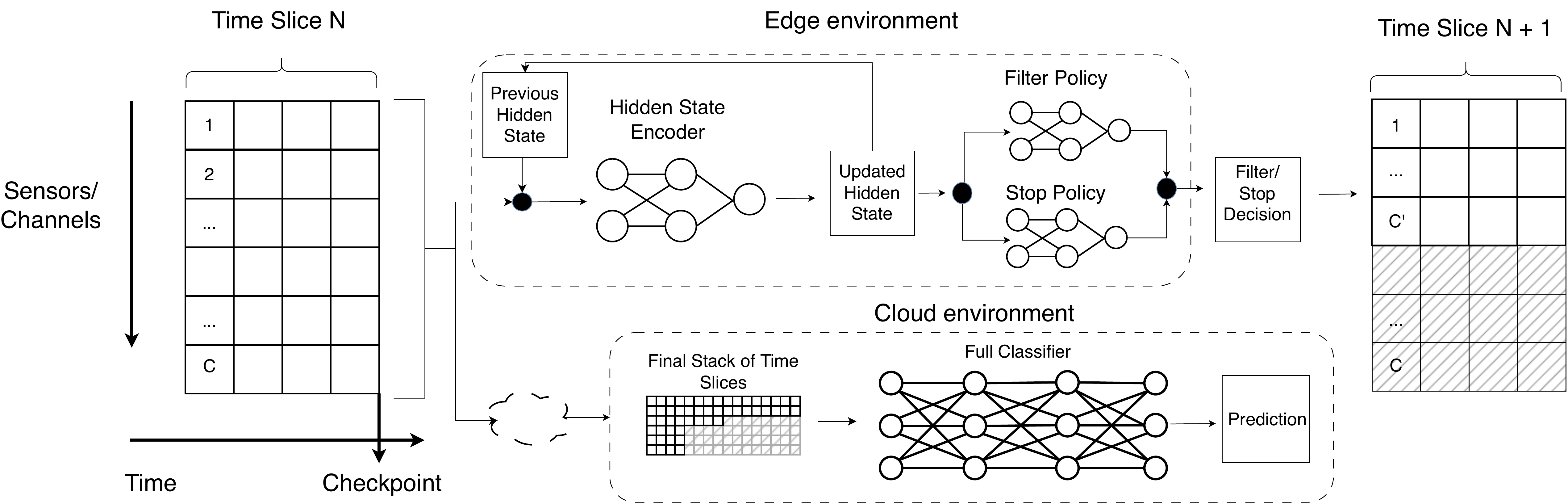}
\caption{Structure of CHARLEE framework}
\label{fig:structure}

\end{figure*}

\section{Proposed framework}

\subsection{Task constraints}
A crucial factor that affects the efficiency of the methods that have to process multivariate time series datasets, either classifiers or feature selectors, is the large dimensionality of both timesteps and channels. As we mentioned above, the current applications of machine learning depend on increasingly larger datasets, aided by the widespread adoption of IoT and sensor data. In theory, for the task of channel adaptation across the time dimension that we are considering, an approach could select, for each timestep $t$ of $T$ in total, any subset of ${0,1,....,C}$ out of C channels to keep, since any channel could have useful information for a class at multiple time points, leading to
\begin{equation}\label{eq:taskcomplexity}
    \bigl(\sum_{i=0}^{C} {C \choose i} \bigr)^{T}
\end{equation} potential paths. We can easily appreciate that even for a small number of channels and timesteps this becomes extremely complex, with the action search space exploding. In addition, even if a reasonable approach was found during training, the computational cost during inference for the channel selection decisions could quickly negate any early-exit benefits in the context where we place our framework.
Thus, to make the task we are considering tractable, without making it unrealistic for real-world use cases, we apply the following constraints:

\textbf{1. Slices across time} Similarly to the existing reinforcement learning approaches, we use a neural network-based policy to decide at each point whether the input processing should stop, filter out some channels or continue. If we would execute this policy at each timestep, as most related work does, there would be a high computational cost associated with it. This could outweigh the energy saved due to the input filtering, without a clear benefit, especially for higher frequency datasets, where a single timestep is unlikely to make a large difference in the classification accuracy. Thus, we decide to split the time dimension into slices of multiple timesteps, with the policy decision taking place at each checkpoint after a slice. This segmentation approach also appears in \cite{li2014early}, albeit with a different rationale and assumptions. This slicing reduces the exponential term of Eq. ~\ref{eq:taskcomplexity} from $T$ to $N$, with $N < T$ (potentially $N \ll T$). We set the checkpoint number as a hyperparameter, and it can be chosen based on domain expertise for a specific use case, or to aim for specific statistics of the input processing savings, such as minimum or maximum percentage. $N$ checkpoints split the time series into $N+1$ slices, so we consider the first checkpoint to be between the first and second time slice, and the last checkpoint to be immediately before the last time slice.

\textbf{2. Channel grouping} Applying the logic of the constraint above to the channel dimension, we can also hypothesize that for datasets with a large number of channels, a very fine-grained selection of single-channel filtering would add complexity to the search space that would be disproportionate to the expected benefit of the early exiting per channel. As a result, we also introduce channel grouping, i.e. simple splitting of the channels into groups that are all filtered out or kept together. We also set this as a hyperparameter, and it reduces the term $C$ of Eq. ~\ref{eq:taskcomplexity} to a term $G$, with $G < C$.

\textbf{3. Fixed filtering order} Another constraint that we set to the task is that there is a fixed ordering of the channels, described in subsection ~\ref{subsection:ordering}, and they are filtered out following this order deterministically. This means that if the policy, at any checkpoint, chooses to keep a subset of K channels, with $K<C$, this subset always contains the same channels, in a given order. The ordering process takes place before the channel grouping and it removes the binomial coefficient of $C$ and $i$ from Eq. ~\ref{eq:taskcomplexity}, leaving a simple summation.

\textbf{4. Monotonically decreasing channels} The last constraint is that the number of channels should decrease monotonically across time, so at each checkpoint, the channels selected to be kept are at most equal to the ones kept up to this checkpoint. This changes the upper limit of the summation term from $C$ to $K_{n-1}$, the channels kept at the previous checkpoint. Although this design choice is not as immediately intuitive as the ones above, it is also motivated by the search space and implementation complexity reduction.

Apart from the theoretical motivation, these constraints also make it possible for us to test the behavior of our framework on a large number of datasets. Given that a single training experiment can take 2-4 hours to complete on our setup, depending on the dataset, and we run experiments for multiple random seeds and parameters, these globally applied restrictions render the research process viable. However, it is important to note that, depending on the training and inference budget, a practitioner can choose to relax some of our suggested constraints, aiming for better performance at the cost of increased training search space or inference cost.

Due to the above constraints, we can describe the early-exiting behavior for a dataset across time with a sequence of numbers. For instance, let us assume a dataset with 4 channels and 3 checkpoints (splitting the dataset into 4 time slices). The utilization vector $\textbf{U} = [1,1,0.5,0.25]$ means that for the first and second time slices, all channels are used. Then, before the 3rd time slice, channels $3$ and $4$ are filtered out, so 50\% of the total channels remain. Finally, before the 4th time slice, channel $2$ is filtered out, so 25\% of the total channels (channel $1$) remain until the end of the series.

\subsection{Channel ranking}\label{subsection:ordering}
In order to satisfy constraints no. 3 and 4, we have to decide which channels to abandon most efficiently, so a ranking of channels according to their utility for the classification task is necessary. We calculate this using the work of Dhariyal et al.\cite{dhariyal2021fast} since it is fast, scalable with the number of channels, and classifier-agnostic, which our framework aims to be. Since we are considering early exiting at different checkpoints per channel, we adopt the following approach during the training phase: For each checkpoint, we truncate the dataset up to it. This means that for the first checkpoint, we consider the first time slice, for the second checkpoint the first two time slices, and so forth in an additive manner. 
We then run the KMeans method from \cite{dhariyal2021fast} on these subsequences of the whole series to generate their rankings. 
Finally, we construct a weighted sum of those rankings, with the weight applied to the first term being 1 and the last a user-defined hyperparameter (we choose 0.1), with a linear reduction for the ones in between. The rationale is that we should abandon first the channels with the highest rankings, which indicates that those channels have the most useful information for the classification task in their initial checkpoints (due to the weighted sum), and thus can be filtered out as time progresses.

\subsection{POMDP formulation}
Similarly to recent works \cite{martinez2020adaptive,hartvigsen2020recurrent,hartvigsen2022stop}, we formulate the channel-adaptive early-exiting problem as a Partially Observable Markov Decision Process (POMDP). The main workflow of the framework, portrayed in Fig. ~\ref{fig:structure}, is as follows: After receiving a time slice of input, at the decision checkpoint, the framework constructs a state $\textbf{S}$, based on this input and the policy filter decisions up to that point. Our reinforcement learning policy observes this state and takes a filtering/stopping action $\textbf{A}$, and a reward $\textbf{R}$ is achieved, reflecting the quality of the action for the final goal of accuracy and savings. The action leads to a new state, $\textbf{S'}$. An episode terminates either because the policy decides to stop (or keep 0\% of the channels) or there are no more time slices. At that point, all the appropriately filtered time slices are passed to the final classifier, which generates the class prediction. A theoretical explanation of the above terms follows below, while the implementation details appear in subsection \ref{subsection:frameworkstruct}.

\textbf{State} At the beginning of each sample processing, the first time slice received consists of all available channels, while the subsequent ones depend on the policy's filtering decisions. Based on these raw input values, an encoder network creates (or updates) a hidden state representation, containing information about the time series content. When a channel has been filtered out, the encoder treats its values as a "mask" value (e.g. 0). The final state $\textbf{S}$ consists of this representation, augmented with the history of filtering actions taken by the agent and the fraction of the checkpoint number over the total checkpoints. We include these additional features to help the policy locate its position in time and take into account its action history to make better decisions about future filtering.

\textbf{Policy}
Our overall policy consists of two parts: 1) A stochastic filtering policy, which decides the percentage $p_{f}$ of total channels to keep for the next time slice based on state $\textbf{S}$, as in $\pi_{\theta_{f}}(S) = p_{f}$ and 2) a stopping policy which decides if the framework will stop the processing completely, based on state $S$ and the output of the filtering policy, as in $\pi_{\theta_{s}}(S, p_{f}) = a_{s}$, where $a_{s}$ is a stopping probability.

\textbf{Action} Regarding the filter policy, for datasets with few channels, there is a limited number of possible filtering actions. A potential implementation could be to have its actions equal to the number of channels + 1 (for the option of keeping 0 channels) and parametrize a multinomial distribution over this set of actions to sample the number of channels to keep. However, for datasets with many channels, the action space expands significantly, and this discrete way of action selection can lead to inefficient exploration during training. We therefore choose to view the problem similarly to other reinforcement learning works with fine-grained discrete action spaces\cite{wu2020intermittent} and treat it as having a continuous action space. We use the filtering policy to generate a float number in the range $[0,1]$ and multiply it by the percentage of channels maintained up to the action checkpoint. The final $p_{f}$ is the nearest float to that result selected from a hyperparameter-based set of numbers. 

To better understand this process, we can examine a specific example of a dataset with 4 channels. We set as a hyperparameter the number of channel groups to be 4, so 1 channel per group. This choice makes the potential filtering actions at the beginning equal to 5 (the policy can keep between 0 and 4 groups), which in turn makes the predefined set of percentage floats [0,0.25,0.5,0.75,1]. At the first checkpoint, the number sampled from the policy distribution is 0.8. We multiply this by the initial percentage (which is always 1) and we select the float closest to the outcome (0.8 $\,\to\,$ 0.75) as the new percentage. As a result, the policy filters out 1 group and keeps the rest (75\% of the total). At the next checkpoint, the number sampled from the policy distribution is 0.4. We multiply it by 0.75 (so we get 0.3), and we select the float closest to the result (0.25). In this case, the policy filters out two more groups and keeps only 1 (25\% of the total).
The stopping policy is not stochastic, but outputs a number in the range $(0,1)$. If the number is over a threshold dependent on the savings factor hyperparameter, the processing terminates.

\textbf{Reward} We want to construct the policy's reward in such a way that it encodes the goals of earliness and accuracy, as multiple related works have considered\cite{mori2017early,martinez2018deep,martinez2020adaptive}. In our framework, the concept of earliness is broader and has to be redefined as input savings, since the goal is to limit the input used for a sample inference, but this can happen both across the time and channel dimensions. As Martinez et al. discuss in \cite{martinez2020adaptive}, we want to give the practitioner the possibility to set their own preferred trade-off between accuracy and savings, depending on the application. We define the hyperparameter $\delta$, which can be set in the range $[0,1]$, with 0 corresponding to maximum accuracy, regardless of savings, and 1 corresponding to maximum savings, with no concern for accuracy.

We formulate the cost of an inference as the sum of the channel utilization percentages across the time slices. So, a sample with a utilization of [1,0.5,0.5,0.25] would have the same cost as one with [1,0.75,0.5,0]. In both cases, the input savings achieved by CHARLEE would be $(4-2.25)/4 = 43.75\%$. Of course, it is trivial to make this sum weighted if a use case requires prioritization of earlier classifications. The minimum inference cost a sample can have is 1 (if the framework exits immediately at the first checkpoint) and the maximum is equal to the number of slices (if the framework does not filter out any of the input). We construct the savings reward by mapping this range from 1 to -1. Regarding accuracy, we set the reward to +1 for a correct classification and -1 for an incorrect one. Thus, at the end of an inference episode, the reward is \begin{equation}\label{eq:reward}
    \mathbf{R} = (1-\delta)*R_{classification} + \delta*R_{savings}
\end{equation}

 \subsection{Framework structure}\label{subsection:frameworkstruct}
Our framework has 3 main modules: (1) The neural network responsible for the creation of a hidden state from the raw time series input; (2) The neural networks implementing our filtering or stopping policy and (3) the final classifier. 

\textbf{1. Hidden state encoder} The first module of the framework, the hidden state encoder, is responsible for mapping the input values of the consecutive time series slices to a fixed hidden state. This hidden state is what is used as input to the policy networks. This approach has been employed in related reinforcement learning methods\cite{hartvigsen2020recurrent,hartvigsen2022stop}, utilizing an RNN. 

The task we are considering presents a particular challenge, i.e. the number of channels is variable across time slices, so the encoder network needs to be able to address this.
Although in theory this issue could be handled by padding the input to the correct shape before passing it to an RNN, we have an additional design goal, in line with the whole philosophy and context of the framework: We would like the encoder to avoid unnecessary computations on the missing channels since they will also have an energy cost on the edge device.

With these requirements in mind, we select a 1-dimensional convolutional network as the encoder, with separate kernels for the separate input channels (or groups of channels). In this manner, when some of the input channels are dropped, the encoder can skip the convolution with the corresponding kernels, similar to some recent dynamic neural network methods\cite{han2021dynamic}. After the convolution, we extract statistical features from the result (max, min, mean, percentage of positive values, mean of positive values and mean of indexes of positive values), and we save those as the hidden state. This achieves two goals: First, it reduces the size of the hidden state, which consequently reduces the computational cost and size of the policy networks. Second, we can correctly update the hidden state based on the current input, by choosing aggregation features that can be globally updated based on their previous value and the current result. This has the additional benefit of giving our solution robustness when different time slice lengths are used since the hidden state will only depend on what timestep each checkpoint is located at and not the intermediate updates. 
Another reason that we select this hidden state representation for the policy network is the success of recent methods that depend on the same extracted features from convolutions for (multivariate) time series classification\cite{dempster2020rocket,tan2022multirocket,pantiskas2022lightwaves}.

\textbf{2. Policy networks} Adopting research on policy gradients in continuous action environments\cite{chou17beta}, the filtering policy is implemented as a feed-forward neural network that outputs two parameters, $\alpha$ and $\beta$, which parametrize a Beta distribution. The stopping policy is not stochastic, but is implemented as a neural network with a sigmoid activation function.

 \textbf{3. Full classifier} The final component of the framework is the full classifier, conceptually located at a remote cloud device, which runs the inference on the appropriately filtered time slices when the input processing is terminated. The classifier has the same requirements as the hidden state encoder, i.e. being able to handle the variability of both the length of the sample and of the channels across time. Although we could build a new classifier that would satisfy those requirements, we wanted to make our framework as classifier-agnostic as possible, to make it easy to apply to existing solutions. Moreover, designing a classifier whose accuracy approaches the existing state of the art is a complex task on its own, and we wanted to validate the performance of our framework in classifiers that are well-known in the community. Thus, we select InceptionTime\cite{ismail2020inceptiontime} and ResNet\cite{wang2017resnet} to experiment with, two deep learning classifiers that have a proven track record in the task of multivariate time series classification\cite{ruizGreatMultivariateTime2021}.
 
 Given this choice of a classifier-agnostic framework, and since both classifiers can handle variable-length inputs, we revert to masking out the filtered channels with a predefined mask value. The rationale of this choice is inspired by the concept of dropout\cite{srivastava14dropout}, where part of the information is zeroed out before being propagated to the next network layer. In addition, this approach has interesting parallels with classifier-agnostic, occlusion-mask-based explainability methods that have been used in computer vision\cite{petsiuk2018rise}, and more recently in the multivariate time series domain\cite{crabbe2021dynamask,mercier2022timereise}.

 \subsection{Training and inference process}

All modules of the framework are trained together end-to-end. The classifier is trained by minimizing the \textbf{cross-entropy loss} $L_{acc}$ between its predicted probabilities and the class labels. For the filter policy network, which involves non-differentiable sampling, we use the REINFORCE method\cite{williams1992simple} to estimate the gradient to update its parameters and maximize the expected reward,
resulting in the loss\cite{schulman2015high}:

{\small
\begin{equation}\label{eq:lfilter}
    L_{filter} = -\mathbb{E} \Bigl[ \sum_{n=0}^{N-1} log \pi_{\theta_{f}}(A_{n}|S_{n})\Bigl[\Bigl(\sum_{j=n}^{N-1}\gamma^{j-n}r_{j}\Bigr) - b(S_{n}) \Bigr]\Bigr]
\end{equation}}
In Eq. ~\ref{eq:lfilter}, $N$ is the number of checkpoints, $A_{n}$ the filter action by the policy $\pi_{\theta_{f}}$, $r$ the reward and $S_{n}$ the state created by the hidden state encoder network, which is co-optimized with the filter policy. Although the reward is calculated once at the end of the classification episode, we propagate it to each checkpoint to train the filtering policy's actions. In addition, as is typical in reinforcement learning, we discount this series of rewards with a (hyperparameter) factor of $\gamma=0.99$, since earlier actions in our formulation of the policy can limit future actions, so they have higher importance. Finally, to reduce the variance of the gradient estimation, we subtract from the reward at each checkpoint a baseline value $b(S_{n})$. This is learned by a network that has the same input as the filter policy and is trained by minimizing the \textbf{mean squared error loss} $L_{baseline}$ between the value and the reward. 

An issue with early classification, that related works also note, is that a model that stops early often during training may never get the opportunity to get information from the full length of the time series samples, and thus make sub-optimal decisions\cite{martinez2020adaptive,hartvigsen2022stop}. Our approach to tackle this issue is that, during training, the action of the stop policy is not taken into account. The processing of a sample can only stop if the filtering action selects 0\% of the channels. Instead, at each checkpoint, we save the result of the classifier on the intermediate filtered input, and the corresponding savings. Thus, we can create an $R_{stop}$ with these two elements, similar to Eq. ~\ref{eq:reward}, for each checkpoint. If at any checkpoint this reward is higher than all future checkpoints, it means that the policy should stop. Thus, the stop policy can be trained by minimizing the \textbf{binary cross-entropy loss} $L_{stop}$ between this binary criterion and the outcome of the policy, without affecting the trajectory during training. A second method we use to address the aforementioned issue is that, for the incorrectly classified samples, we add to $L_{acc}$ the \textbf{cross-entropy loss} $L_{full}$ between the classifier result on the full, unfiltered sample and its label. In this way, even if the policy does not observe the whole input, the classifier gets the chance to be trained on it, potentially improving the reward for future iterations. Summarizing, the total loss of the framework is:
\begin{equation}
    L_{total} = L_{acc} + L_{full} + L_{filter}  + L_{baseline} + L_{stop}
\end{equation}
\section{Experiments}
We want to verify the expected behavior of our framework, and we focus on assessing if its flexibility can indeed add value to the early classification paradigm, by achieving better accuracy for equal savings, for at least some of the real datasets. To this end, for both the synthetic and the real datasets, we follow \cite{martinez2020adaptive} and we use as comparison the performance of the classifiers with static exit points across the time dimension. These exit points are dictated by the savings achieved by CHARLEE. For instance, if for a dataset and a given $\delta$ CHARLEE achieves average input savings of 20\%, the point of reference would be the performance of the classifier on the dataset truncated at 80\% of its length. We term this point of reference Time-Only Early Exiting (TOEE). 

\subsection{Datasets}
\subsubsection{Synthetic dataset}
Since there is no ground truth for the optimal exit times for each channel for the public multivariate datasets, we construct a multivariate synthetic dataset in order to validate our framework. Instead of introducing completely artificial signals, we use samples from the univariate dataset ElectricDevices of the UCR dataset collection\cite{dau2019ucr}, so the synthetic problem is closer to a real-world one. We choose this dataset because the signals per class are clear and simple enough to be human-understandable. We create a dataset of 4 channels and 96 timesteps, intended to be split in 4 time slices. We select characteristic class prototypes from the ED dataset, and we preprocess them appropriately so that their discriminative parts, such as peaks, are distributed across channels and time slices to create the following 8 classes:
\begin{itemize}
\item Classes 1,2: The unique combination of signals in the 1st time slice makes it so that a perfect early-exit classifier would stop immediately after it for these samples. The ideal utilization is [1,0,0,0] and the ideal input savings would be 75\%.
\item Classes 3,4: The signal that discriminates these samples is at the 4th slice of the 4th channel, so the ideal utilization would be [1,0.25,0.25,0.25] and the ideal input savings would be 56.25\%.
\item Classes 5,6: The signals that discriminate these samples are at the 4th slices of the 3rd and 4th channels. However, the 2nd slice is similar to that of classes 7,8, so a perfect early-exit classifier could not exit before that. The ideal utilization would be [1,1,0.5,0.5] and the ideal input savings would be 25\%.
\item Classes 7,8: The signals that discriminate these samples are at the 4th slice of all channels, so the ideal utilization would be [1,1,1,1] and no input saving is possible.
\end{itemize}

\begin{figure}[tp]
\centering
\includegraphics[width=\linewidth]{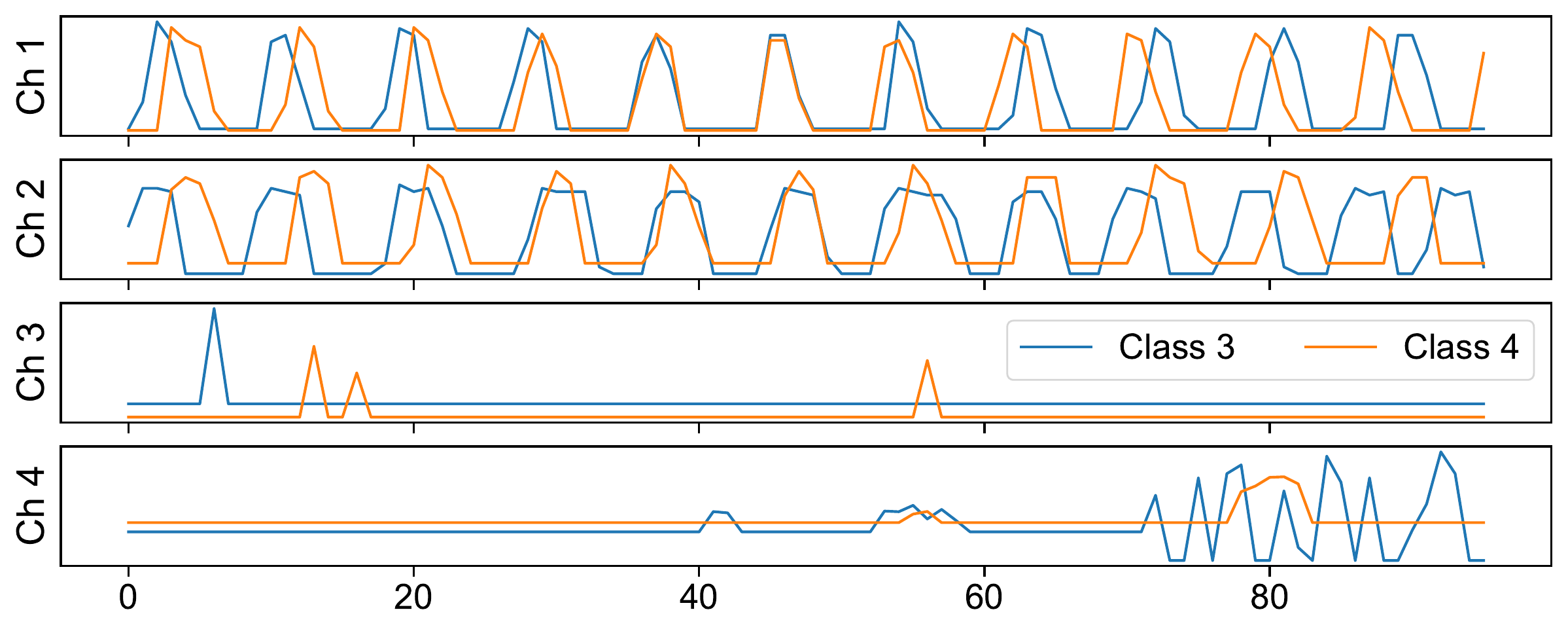}
\caption{Samples of classes 3 and 4, showcasing the logic of the synthetic dataset}
             \label{fig:classexamples}
\end{figure}

An example of classes 3 and 4 is presented in Fig. ~\ref{fig:classexamples}, demonstrating the format of the dataset. By comparing the stopping points of our framework with the ideal stopping points, we can draw insights about the behavior of the reinforcement learning policy.

\subsubsection{Real datasets}

For benchmarking, we first start with the initial set of the UEA collection of multivariate datasets \cite{bagnallUEAMultivariateTime2018}. In addition, we use the following machinery-fault related datasets: \textbf{MAFAULDA} (Machinery Fault Database) \cite{MAFAULDAMachineryFault}, a dataset with 8 sensor measurements on a machine fault simulator, and \textbf{CWRU} \cite{CWRU}, a bearing fault dataset from Case Western Reserve University. In order to fairly assess the performance of our framework, we want to make sure that any potential input savings are due to the learned policy, and not because of the dataset format, so we apply the following process: First, we get baseline test accuracies for both deep learning models for each full dataset. Then, we limit each dataset to 5,10,...,95\% of its original length and train the models on each of these subsequences. If the test accuracy approaches the baseline accuracy at any of these points, we keep the truncated version of the dataset to experiment on, since it means that the full length is not necessary by default for classification. In addition, we exclude datasets that demonstrate irregular (for the purpose of early classification) behavior during this process (e.g. oscillating accuracy). 

After this filtering process, 26 datasets remain for experimentation, covering a large spectrum of use cases and with a large variety in channel and timestep dimensions. Although we do not expect all of them to be suitable for the successful application of our framework, this extensive testing gives us the opportunity to observe its strengths and points of improvement for future work, as well as present its performance in a fair manner, without only selecting favorable results.
\subsection{Experimental setup}
All experiments were run on the DAS-6 infrastructure \cite{balDAS2016}, on nodes with 24-core AMD EPYC-2 (Rome) 7402P CPUs, NVIDIA A6000 GPUs, and 128 GB of RAM. We implement the framework using PyTorch\cite{paszke2019pytorch}, and we use the implementations of the tsai package\cite{tsai} for InceptionTime and ResNet. We repeat all experiments 5 times with different random seeds, and we present the averages of the metrics. We use 20\% of each train set as a validation set, and we use the weights that result in the best validation score for testing. 

We use a common number of checkpoints (4) for all datasets, and we set the number of channel groups equal to the minimum between the number of channels of each dataset and 10. These choices are used to get an estimation of the overall performance of the framework and are unlikely to produce optimal results for all datasets, since each use case is unique and requires domain expertise and/or hyperparameter search to find better configurations. However, this detailed dataset exploration is out of scope for this work and is best suited for applied future work.

\section{Results}
\subsection{Synthetic dataset}

\begin{figure}[tp]
\centering
\includegraphics[width=\linewidth]{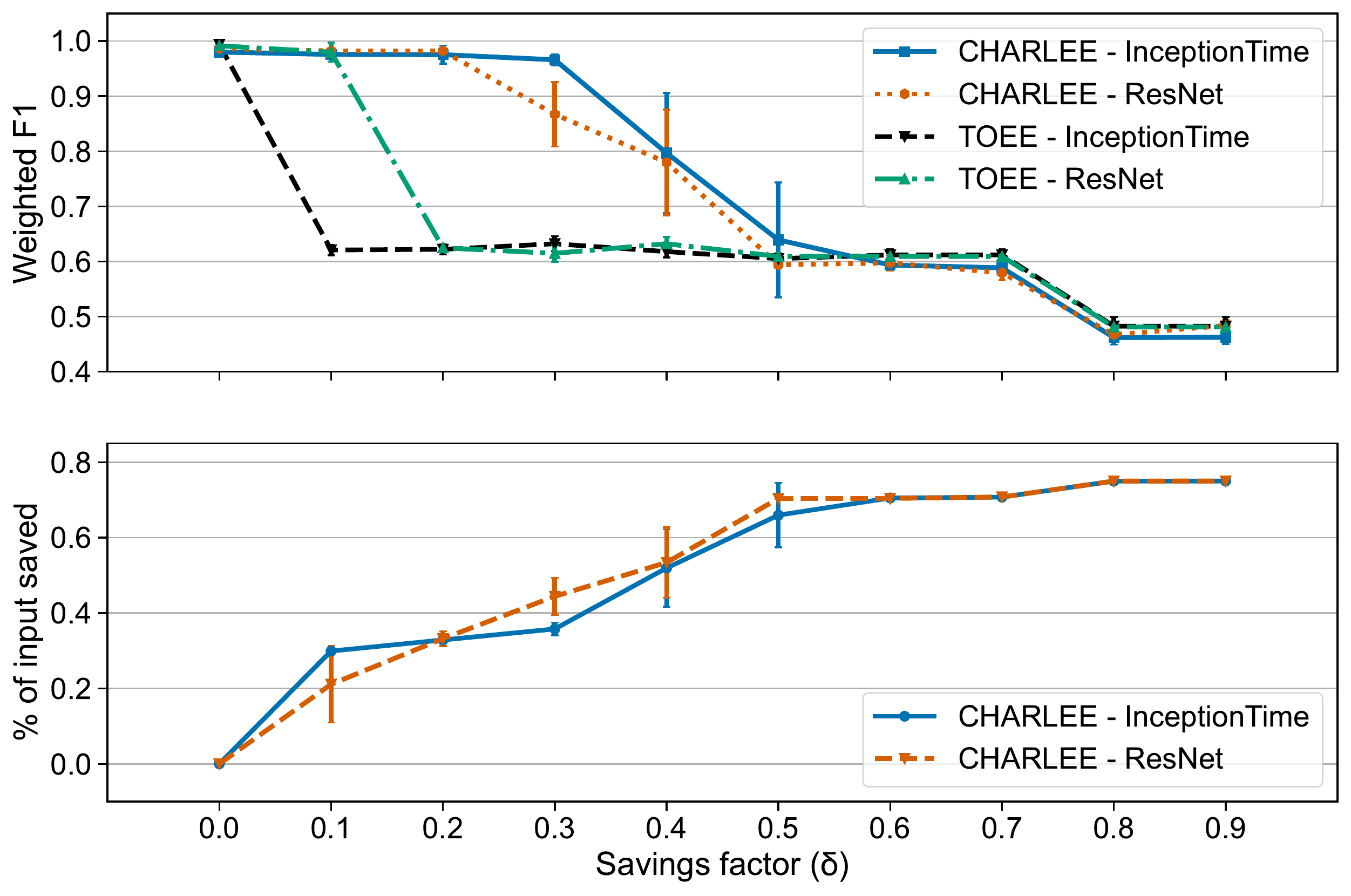}
\caption{Comprarison of CHARLEE and TOEE for the utilized DL models at various earliness factors. The input savings achieved by CHARLEE are used as reference to truncate the dataset used by the TOEE}
             \label{fig:synthresults}
\end{figure}

We can see in Fig. ~\ref{fig:synthresults} the F1 and input savings achieved by CHARLEE, as well as the F1 achieved by the Time-Only Early Exiting for equal input savings across time. Firstly, it is clear that CHARLEE displays the expected behavior, with F1 decreasing overall and input savings increasing as the savings factor increases. In addition, for low savings factors, CHARLEE can maintain almost perfect F1 while saving a significant percentage of the input. Specifically, for $\delta=0.2$, CHARLEE saves more than 30\% of the input for both DL models. For InceptionTime, at $\delta=0.3$, CHARLEE approaches the ideal 40\% in savings across the synthetic dataset, while keeping F1 close to 1. In contrast to that, for savings of more than 25\%, the TOEE models following the traditional early classification paradigm experience a significant drop in F1, since they start missing the information contained in the last slice of the synthetic dataset, even as they maintain all channels. As the savings factor increases, the performances of CHARLEE and TOEE start to converge again, as the policy is trying to achieve maximum permissible savings, and its flexibility is reduced.

\subsection{Real datasets}

We have discussed that the globally-applied hyperparameters are unlikely to result in optimal performance of CHARLEE for all datasets. Moreover, not all multivariate time series patterns can be suitable for the application of CHARLEE in the way we have formulated its objective. Therefore, we expect that we will notice an advantage over the early classification paradigm in only a subset of all tested datasets. In Table ~\ref{table:winslosses}, we can see the number of datasets where CHARLEE indeed offers flexibility to the TOEE for the real datasets and the ones where TOEE is preferable, for $\delta$ up to 0.4. We skip $\delta=0$ since it is trivial to just use a normal classifier and we do the same for higher $\delta$ factors, guided by the results on the synthetic dataset, as we see that CHARLEE rewards savings more, and its performance in accuracy drops. We consider a method to perform better on a dataset if the mean of its F1 scores across runs is more than 0.01 (1 percentage point) higher than that of the alternative.
\begin{table}[tp]
\caption{Number of datasets where CHARLEE is preferable over TOEE with equal savings, or vice versa, for different values of $\delta$}\label{table:winslosses}
\begin{center}
\begin{tabular}{|c|ccc|ccc|}
\hline
      & \multicolumn{3}{c|}{InceptionTime}                   & \multicolumn{3}{c|}{ResNet}                           \\ \hline
Delta & \multicolumn{1}{c|}{CHARLEE} & \multicolumn{1}{c|}{TOEE}  & Ties & \multicolumn{1}{c|}{CHARLEE}  & \multicolumn{1}{c|}{TOEE}  & Ties \\ \hline
0.1   & \multicolumn{1}{c|}{8} & \multicolumn{1}{c|}{11} & 7 & \multicolumn{1}{c|}{10} & \multicolumn{1}{c|}{11} & 5 \\ \hline
0.2   & \multicolumn{1}{c|}{8} & \multicolumn{1}{c|}{12} & 6 & \multicolumn{1}{c|}{7}  & \multicolumn{1}{c|}{13} & 6 \\ \hline
0.3   & \multicolumn{1}{c|}{6} & \multicolumn{1}{c|}{15} & 5 & \multicolumn{1}{c|}{7}  & \multicolumn{1}{c|}{16} & 3 \\ \hline
0.4   & \multicolumn{1}{c|}{6} & \multicolumn{1}{c|}{16} & 4 & \multicolumn{1}{c|}{7}  & \multicolumn{1}{c|}{15} & 4 \\ \hline
\end{tabular}
\end{center}
\end{table}

\begin{table}[tp]
\caption{F1 achieved by CHARLEE for $\delta=0.1$ compared to TOEE approach with equal savings}\label{table:ef1}
\begin{center}
\begin{tabular}{|c|cc|cc|}
\hline
Dataset & \multicolumn{2}{c|}{InceptionTime}                   & \multicolumn{2}{c|}{ResNet}                          \\ \hline
        & \multicolumn{1}{c|}{CHARLEE}        & TOEE             & \multicolumn{1}{c|}{CHARLEE}        & TOEE             \\ \hline
NTPS    & \multicolumn{1}{c|}{\textbf{0.808}} & 0.684          & \multicolumn{1}{c|}{\textbf{0.823}} & 0.743          \\ \hline
EW      & \multicolumn{1}{c|}{\textbf{0.629}} & 0.553          & \multicolumn{1}{c|}{\textbf{0.7}}   & 0.631          \\ \hline
LSST    & \multicolumn{1}{c|}{\textbf{0.568}} & 0.507          & \multicolumn{1}{c|}{\textbf{0.642}} & 0.571          \\ \hline
EMOP    & \multicolumn{1}{c|}{\textbf{0.725}} & 0.665          & \multicolumn{1}{c|}{\textbf{0.725}} & 0.66           \\ \hline
BM      & \multicolumn{1}{c|}{\textbf{0.694}} & 0.645          & \multicolumn{1}{c|}{\textbf{0.765}} & 0.629          \\ \hline
FD      & \multicolumn{1}{c|}{\textbf{0.626}} & 0.582          & \multicolumn{1}{c|}{\textbf{0.599}} & 0.563          \\ \hline
SAD     & \multicolumn{1}{c|}{\textbf{0.933}} & 0.917          & \multicolumn{1}{c|}{\textbf{0.924}} & 0.899          \\ \hline
DDG     & \multicolumn{1}{c|}{\textbf{0.483}} & 0.468          & \multicolumn{1}{c|}{0.427}          & \textbf{0.479} \\ \hline
MAF     & \multicolumn{1}{c|}{0.944}          & 0.937          & \multicolumn{1}{c|}{0.937}          & \textbf{0.952} \\ \hline
MIND    & \multicolumn{1}{c|}{0.391}          & 0.389          & \multicolumn{1}{c|}{0.367}          & \textbf{0.39}  \\ \hline
PS      & \multicolumn{1}{c|}{0.123}          & 0.124          & \multicolumn{1}{c|}{\textbf{0.241}} & 0.191          \\ \hline
EP      & \multicolumn{1}{c|}{0.865}          & 0.867          & \multicolumn{1}{c|}{\textbf{0.888}} & 0.875          \\ \hline
HB      & \multicolumn{1}{c|}{0.674}          & 0.679          & \multicolumn{1}{c|}{0.67}           & 0.667          \\ \hline
CHAR    & \multicolumn{1}{c|}{0.975}          & 0.98           & \multicolumn{1}{c|}{0.967}          & \textbf{0.98}  \\ \hline
CWRU    & \multicolumn{1}{c|}{0.971}          & 0.98           & \multicolumn{1}{c|}{0.967}          & 0.965          \\ \hline
LIB     & \multicolumn{1}{c|}{0.793}          & \textbf{0.805} & \multicolumn{1}{c|}{0.822}          & \textbf{0.888} \\ \hline
RS      & \multicolumn{1}{c|}{0.738}          & \textbf{0.758} & \multicolumn{1}{c|}{\textbf{0.798}} & 0.763          \\ \hline
MHAR    & \multicolumn{1}{c|}{0.916}          & \textbf{0.937} & \multicolumn{1}{c|}{0.926}          & 0.926          \\ \hline
HW      & \multicolumn{1}{c|}{0.339}          & \textbf{0.365} & \multicolumn{1}{c|}{0.21}           & \textbf{0.273} \\ \hline
AWR     & \multicolumn{1}{c|}{0.862}          & \textbf{0.897} & \multicolumn{1}{c|}{0.884}          & 0.89           \\ \hline
CR      & \multicolumn{1}{c|}{0.893}          & \textbf{0.941} & \multicolumn{1}{c|}{0.867}          & \textbf{0.924} \\ \hline
UW      & \multicolumn{1}{c|}{0.793}          & \textbf{0.861} & \multicolumn{1}{c|}{0.722}          & \textbf{0.772} \\ \hline
HMD     & \multicolumn{1}{c|}{0.237}          & \textbf{0.309} & \multicolumn{1}{c|}{0.227}          & \textbf{0.246} \\ \hline
ER      & \multicolumn{1}{c|}{0.611}          & \textbf{0.703} & \multicolumn{1}{c|}{0.694}          & 0.697          \\ \hline
SRS1    & \multicolumn{1}{c|}{0.696}          & \textbf{0.801} & \multicolumn{1}{c|}{0.745}          & \textbf{0.804} \\ \hline
JV      & \multicolumn{1}{c|}{0.817}          & \textbf{0.93}  & \multicolumn{1}{c|}{0.901}          & \textbf{0.915} \\ \hline
\end{tabular}
\end{center}
\end{table}

We see that CHARLEE manages to outperform the early classification paradigm in multiple datasets for $\delta=0.1$, whereas TOEE is preferable in others, as we expected. As $\delta$ increases, the number of cases where CHARLEE performs better decreases, following the pattern we notice on the synthetic dataset. In Table ~\ref{table:ef1}, we can see in more detail the achieved F1 score for $\delta = 0.1$, for CHARLEE and TOEE, sorted by best to worst performance for InceptionTime. The bold numbers indicate where each method performs better. We notice that there is a considerable overlap of the behavior of CHARLEE between InceptionTime and ResNet, confirming that its performance mainly depends on the data dynamics of each dataset, given our problem formulation. At this point, we should mention that the TOEE paradigm has an advantage, as point of reference, over CHARLEE. This is because, although the savings are the same, the dataset for the TOEE method is truncated at a timestep granularity, while CHARLEE operates on coarse-grained time slices. Even so, as we established, CHARLEE manages to prove its worth in several datasets.

There are additional possibilities to improve the performance of CHARLEE. We have already mentioned the relaxation of the problem constraints. Moreover, although our framework is classifier-agnostic, the structure of the classifier models could be improved with modules that can handle the variability of the filtered input in a more efficient manner and result in better accuracy, such as \cite{sawada2022convolutional}. Finally, another potential point of improvement, given that reinforcement learning algorithms can be significantly affected by hyperparameters\cite{islam2017reproducibility}, would be a fine-tuning of components such as the discount factor $\gamma$, the scale of the rewards, or the ranking method of the channels, based on a specific dataset.

\section{Conclusion}
In this work we present CHARLEE, a framework for channel-adaptive, reinforcement learning-based early exiting during multivariate time series classification. Our framework combines the concept of early classification across time with filtering out channels that are no longer useful, thus presenting a more flexible, fine-grained approach to the problem of early classification, adapted for the domain of multivariate time series. The main purpose of the framework is to reduce the input necessary for a correct classification of a sample in a more efficient way than simple early exiting across time. The context in which we present our framework is edge intelligence, where this input reduction limits the data that have to be transferred both from the sensors to the edge device, and from the edge device to the cloud classifier. This could in turn translate to energy savings for the edge device and the sensors, where it is most critical.

We verify the behavior of our framework on a synthetic dataset and we test it extensively on a large collection of multivariate datasets. We show that the accuracy difference compared to a theoretical, time-only early classifier with equivalent savings depends on the dataset dynamics, but its added flexibility logic shows promising results, especially considering the potential improvement after hyperparameter fine-tuning on a use-case basis. 

In summary, we have shown that for many (though not all) datasets it is already advantageous to choose CHARLEE over time-only early exiting. An inviting direction for future research is trying different policy gradient methods for the same channel-adaptive early-exiting task, such as the ones described in \cite{duan2016benchmarking}, which can avoid some of the pitfalls of REINFORCE and give more robust results.
Finally, it would be interesting to explore more the similarities with the explainability methods for black-box models, e.g. by adapting the framework to encourage it to detect the different data patterns for the separate classes, expanding on recent work on the subject\cite{gao2022rlpattern}.

\section*{Acknowledgment}

We would like to thank Vincent François, Floris den Hengst, and Jacob Kooi for the insightful discussions and invaluable feedback on the reinforcement learning concepts.

\bibliographystyle{IEEEtran}
\bibliography{IEEEabrv,references}

\end{document}